\documentclass[twocolumn,9pt]{article}
\usepackage{flushend}
\usepackage{spconf,amsmath,graphicx}
\usepackage{subfigure}
\usepackage{multirow}
\usepackage{longtable}
\usepackage{amsthm}
\usepackage{amssymb }
\usepackage{amsmath}
\usepackage{bm}
\usepackage{mathrsfs}
\usepackage{amsfonts}
\usepackage{longtable}
\usepackage{multirow}
\usepackage{algorithm, algpseudocode}
\usepackage{cite}
\usepackage{color}

\def\ij{_{ij}}
\def\1{^{(1)}}

\def\NIC{N_{l}}
\def\NIPC{P_1}
\def\NICC{R}
\def\NPC{N_1}
\def\NCC{M}

\def\i{_{i}}
\def\j{_{j}}

\def\s{\bm{s}}
\def\v{\bm{v}}

\def\u{\bm{u}}

\def\W{\bm{W}}

\algnewcommand\Input{\item[\hspace{6pt}\textbf{Input:}]}
\algnewcommand\Output{\item[\hspace{6pt}\textbf{Output:}]}
\algnewcommand\OutputVal{\textbf{output} }

\title{Improved Explainability of Capsule Networks: Relevance Path by Agreement}
\name{Atefeh Shahroudnejad$^\dagger$,  Arash Mohammadi$^\dagger$, and Konstantinos N. Plataniotis$^\ddagger$}
\address{$~^\dagger$Concordia Institute for Information Systems Engineering,  Concordia University, Montreal, QC, Canada \\
$~^\ddagger$Department of Electrical and Computer Engineering, University of Toronto, Toronto, ON, Canada\\
Emails: $\{$a\underline{\space}hahrou, arashmoh$\}$@encs.concordia.ca; kostas@ece.utoronto.ca
 \thanks{This work was partially supported by the Natural Sciences and Engineering Research Council (NSERC) of Canada through the NSERC Discovery Grant RGPIN-2016-049988.}}
\begin{document}
\ninept
\maketitle
\begin{abstract}
Recent advancements in signal processing and machine learning domains have resulted in an extensive surge of interest in deep learning models due to their unprecedented performance and high accuracy for different and challenging problems of significant engineering importance. However, when such deep learning architectures are utilized for making critical decisions such as the ones that involve human lives (e.g., in medical applications), it is of paramount importance to understand, trust, and in one word ``explain" the rational behind deep models' decisions. Currently, deep learning models are typically considered as black-box systems, which do not provide any clue on their internal processing actions. Although some recent efforts have been initiated to explain behavior and decisions of deep networks, explainable artificial intelligence (XAI) domain is still in its infancy. In this regard, we consider capsule networks (referred to as CapsNets), which are novel deep structures; recently proposed as an alternative counterpart to convolutional neural networks (CNNs), and posed to change the future of machine intelligence. In this paper, we investigate and analyze structures and behaviors of the CapsNets and illustrate potential explainability properties of such networks. Furthermore, we show possibility of transforming deep learning architectures in to transparent networks via incorporation of capsules in different layers instead of convolution layers of the CNNs.
\end{abstract}
\section{Introduction} \label{sec:Introduction}

Nowadays, advanced machine learning techniques~\cite{Bishop:2006} have encompassed all aspects of human life including complicated tasks. As such, several critical decisions are now made based on predictions provided by machine learning models without any human supervision or participation. It is, therefore, of paramount importance to be able to trust a model, validate its predictions, and make sure that it performs completely well on unseen or unfamiliar real world data. For example, in critical domains such as medical applications~\cite{medical:2016} or self-driving cars~\cite{car:2016}, even a single incorrect decision is not acceptable and could possibly lead to a catastrophic result. For guaranteeing reliability of a machine learning model, it is significantly important to understand and analyze rational reasons behind the decisions made by such sophisticated and advanced models,  in other words, we need to be able to  open the black-box.
On the other hand, deep models~\cite{deep:2016} are considered as one of the most successful methods in  several areas especially in image processing and computer vision domains. However, as deep architectures become more complex and introduce more nonlinearity, their structures become less transparent and it is harder to understand what operations or input information lead to a specific decision. Moreover, in scenarios where sufficient data for training is not available, which is a common case when it comes to training deep neural networks such as Convolutional Neural Nets (CNNs)~\cite{ImgN:2014}, probability of error will increase drastically, which further necessitates an urgent need for interpretation of machine learning architectures.

\begin{figure}[t!]
\centering
\includegraphics[width=0.5\textwidth]{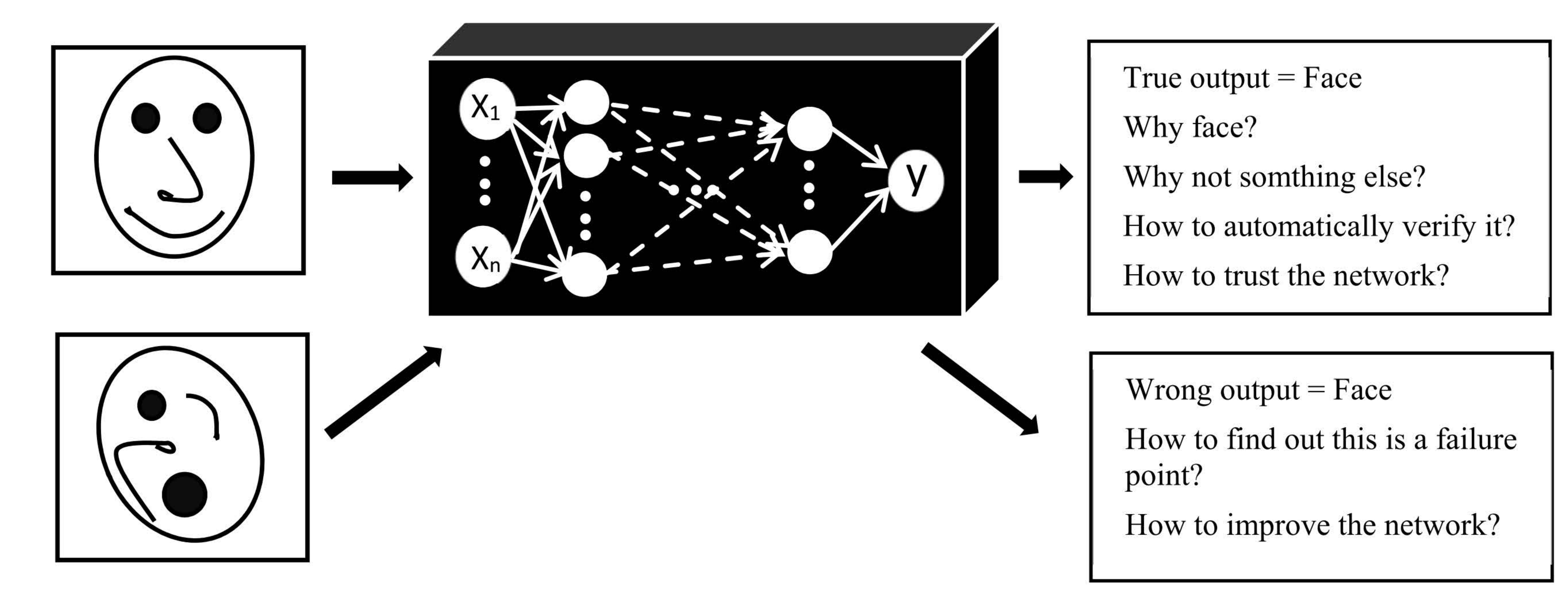}
\caption{\footnotesize Black-box deep neural networks and explainability concept. Feeding an input image and getting a prediction from the other side.}
\label{fig:black}
\vspace{-.3in}
\end{figure}
Explaining a model means providing extra qualitative information about why the model reach to a specific decision regarding the input components' relationship (e.g., different patches of an image~\cite{Marco:2016}). Explanation can be considered as opening incomprehensible black-box and seeing inside. In other words, the main goal is to find answers to questions of like: What is happening inside a neural network? What does each layer of a deep architecture do? What features a deep network is looking for? Fig.~\ref{fig:black} illustrates a graphical representation of this black-box concept, where the input is an image and the network prediction would be a single word (e.g., face or cat). As can be seen, such a single output provides no evidence for confirming the truth of predictions or rejecting incorrect predictions without having access to the ground-truth. Main advantages of using an explainable model are as follows: Verification of the model; Improving it by understanding failure points; Extracting new insights and hidden laws of the model, and; Finally, identifying modules responsible for incorrect decisions~\cite{Samek:2017}.

Explainability can be fulfilled visually, text based, example based, and/or by
relating inputs and learned parameters~\cite{Lipton:2016}. In this regard, recently there has been a great surge of interest~\cite{Kahng:2018, SA:2010, LRP:2015, koh:2017, Kuo:2018, Guo:2017, Samek:2018, Samek:2017-2,sail:2012, Selv:2016, Huang:2017} for development of methodologies to make neural network models explainable. For example, an early work in this area is named sensitivity analysis (SA)~\cite{SA:2010}, which measures how much changing each pixel effects the final decision and then a heatmap is  obtained  referred to as the explainability feature. But this heatmap does not show which patches or pixels play more important role in the decision making process. Moreover, it more looks like a saliency map~\cite{sail:2012}, which uses unique frequencies or focus points or other features to find regions of  interest, however, pixels identified as salient regions are not necessarily the pixels being involved in making the predictions.
In layer-wise relevance propagation method (LRP)~\cite{LRP:2015}, each prediction is decomposed by redistributing backward through the network's layers using redistribution rules and finding a relevance path. Relevance of each neuron or pixel indicates how much it contributes to the decision. The LRP approach is unsatisfying when we have a more complex or nested architecture. Reference~\cite{Marco:2016}, proposed Local Interpretable Model-agnostic Explanations (LIME) approach, which explains the prediction by approximating the original model with an interpretable model around several local neighbourhoods. Reference~\cite{koh:2017} proposed to identify the most responsible training points by tracing back the prediction using influence functions,  which estimate effects of changing each training point on that prediction. Gradient-weighted Class Activation Mapping (Grad-CAM)~\cite{Selv:2016} approach is another explanation method for CNNs which uses gradient to obtain localization map as a visual explanation and finds important layers for each class. Finally, Reference~\cite{Huang:2017} proposed a verification framework for increasing trustworthy and explainability of deep networks. It assumes that there is a region in each input which specifically determines its class category and if the prediction shows this category, its input should include the saliency points of that region.

Although recently different research works are developed for explaining complex behaviour of  deep neural networks, especially for visual tasks as briefly outlined above, explainable artificial intelligence (XAI) is still in its infancy and needs significant research to further open the black-box. In this regard, we focus on a very recently proposed deep network architecture referred to as Capsule Networks (CapsNets)~\cite{Caps:2017}, which is a turning point in the deep learning research. In particular, we investigate CapsNets' architecture and  behavior from explanability perspective. Through analyzing different underlying components of the CapsNet architecture and its learning mechanism (routing-by-agreement), we illustrate that this revolutionized deep net has more intrinsic explainability properties. In particular, we show that CapsNets automatically form the verification framework introduced in~\cite{Huang:2017}, learn regions of interest which determine class category, and as such improve trustworthy and explainability of deep networks. Besides, CapsNets inherently create the relevancy path (introduced in~\cite{Marco:2016}) and we refer to it as relevance by agreement as a bi-product of the routing-by-agreement algorithm, which adds another level of explanability to this architecture.

The rest of this paper is organized as follows: Section~\ref{sec:framework} reviews CapsNets.  Section~\ref{sec:WTE} investigates CapsNets explainability properties. Section~\ref{sec:EXP} shows some explanations of CapsNets on MNIST dataset. Finally, Section~\ref{sec:con} concludes the paper.
\section{CapsNets Problem Formulation} \label{sec:framework}
\begin{figure}[t!]
\centering
\includegraphics[width=0.3\textwidth]{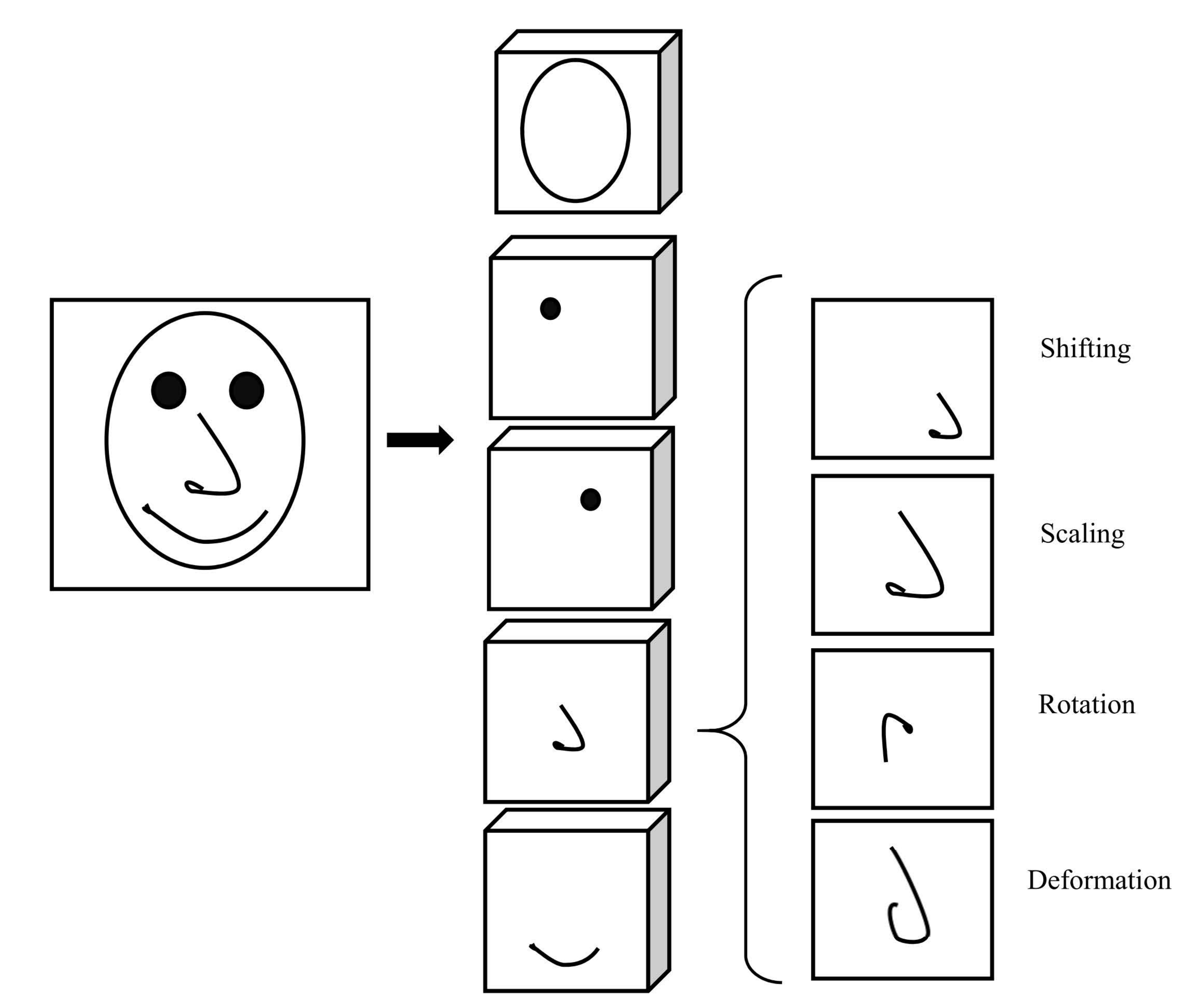}
\caption{Examples of instantiation parameters for a nose capsule in face detection problem. }
\label{fig:trans}
\vspace{-.2in}
\end{figure}

Generally speaking, CNNs learn to classify objects using convolutional operations by first extracting low level features in initial layers and then stacking up the preliminary features learned through initial layers to extract more complex features in higher layers. However, CNN architecture ignores  the hierarchy between layers (as is present in the human brain), which limits their modeling capabilities. CNNs try to overcome (mask) such limitations by utilization of a large amount of training data. The first problem (limitation) is that CNNs use sub-sampling in pooling steps to transfer more important features to the next layer. Therefore, some viewpoint changes and precise spatial relationships between higher level components will be lost. The other problem is that CNNs are not robust against new viewpoints, because they can not extrapolate their geometric information.
For these reasons, CapsNet has been proposed to replace invariance concept with equivalence, which means that if the input is spatially transformed, the network adapts accordingly and responses properly. A capsule is a group of neurons nested inside a layer and is considered as the main building block of the CapsNet architecture. Actually, the idea behind introducing a capsule is to encapsulate (possibly large) number of pose information (e.g., position, orientation, scaling, and skewness) together with other instantiation parameters (such as color and texture) for different parts or fragments of an object. This multilayer structure is deep in width instead of being deep in height. It can be considered as a parse tree because each active capsule chooses a capsule in the next layer as its parent in the tree. By incorporation of capsules instead of neurons and deepening the width, therefore, CapsNets can better handle different visual stimulus and provide a better translational invariance compared to pooling methods (max/average pooling) used in CNNs. Fig.~\ref{fig:trans} shows a nose CapsNet for face detection problem.  In this  illustrative example, there are 5 possible facial component capsules (we note that these are being automatically formed by the CapsNet) and 4 instantiation parameters (i.e., shifting, scaling, rotation, and deformation), which are also extracted automatically by the network.~It is worth mentioning that in designing a CapsNet architecture, we~only specify the number of capsules and the number of instantiation parameters per capsule, the network then learns specifics automatically.

\begin{figure}[!t]
\centering
\includegraphics[width=0.45\textwidth]{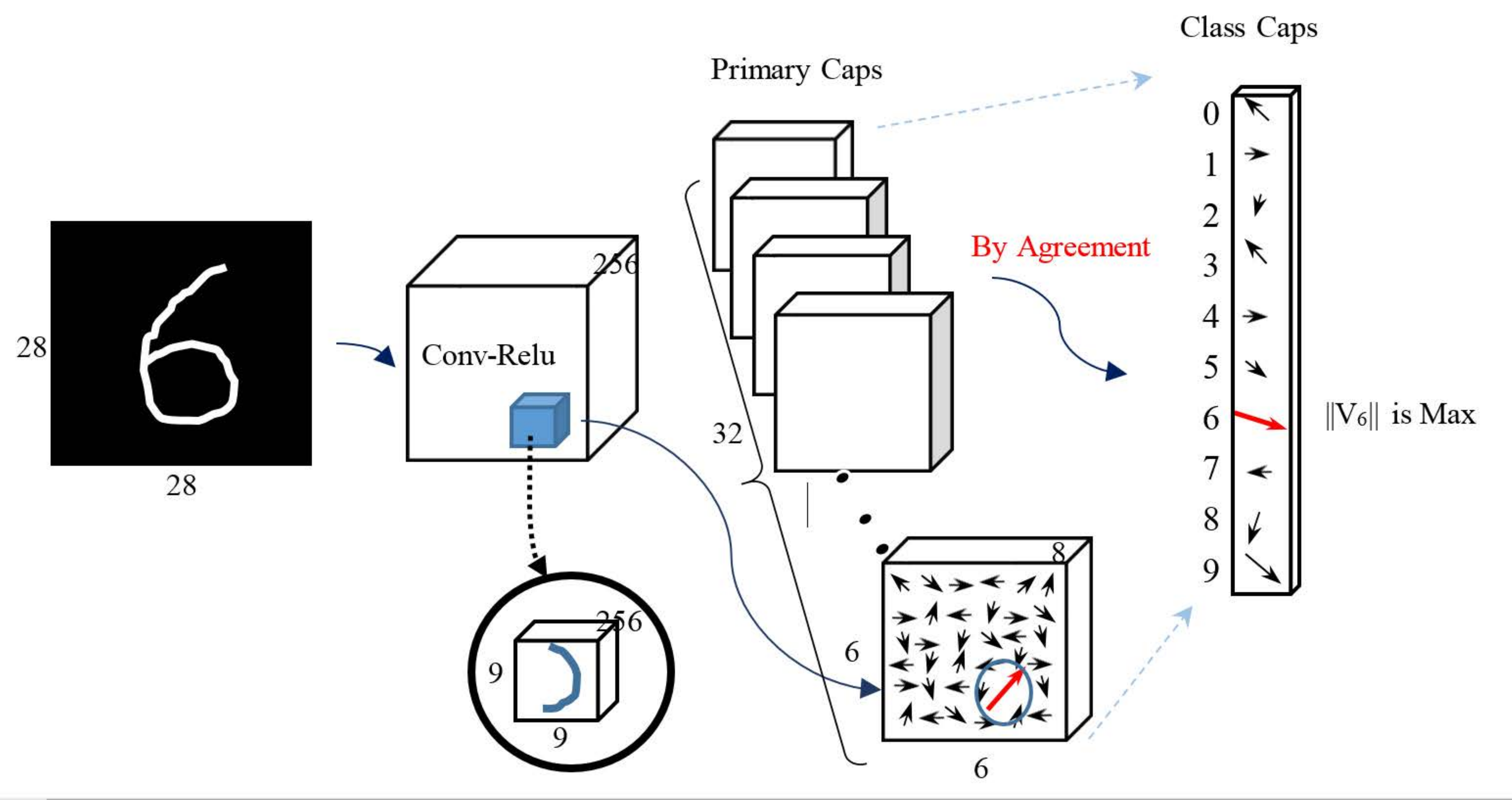}
\caption{\footnotesize Detection architecture for a three layers CapsNet. Each arrow in the Primary-Caps layer indicates activity vector of a capsule. The red arrow shows an activated capsule with higher magnitude for the example kernel introduced to find right curve fragment of digit $6$. }
\label{fig:arch}
\vspace{-.2in}
\end{figure}

Fig.~\ref{fig:arch} shows detection architecture for a CapsNet with the following three layers: (i) Convolutional layers; (ii) A \textbf{\textit{Primary-Capsule~(PC)}} layer (the first capsule layer of a CapsNet architecture) consisting of $\NPC$ capsules, and; (iii) A \textbf{\textit{Class-Capsule~(CC)}} layer (the last capsule layer of a CapsNet architecture) with $\NCC$ capsules.
For simplicity of the presentation, first we further describe the  three layers CapsNet shown in Fig.~\ref{fig:arch}.
In this architecture, first the input is fed into the  convolutional layers for extracting local features from the pixels. The next layer is the PC layer, where each capsule $i$, for ($1 \leq i \leq \NPC$),  has an activity vector ($\u_i \in \mathbb{R}^{\NIPC}$)  to encode $\NIPC$ spatial information (instantiation parameters). Capsules in the PC layer can be grouped into several blocks referred to as ``Component Capsules''. In the next step, the output vector $\u_i$ of the $i^{\text{th}}$ PC, for ($1 \leq i \leq \NPC$), is fed into the $j^{\text{th}}$ CC, for ($1 \leq j \leq \NCC$), using weight matrix $\W\ij \ \in\mathbb{R}^{(\NICC\times\NIPC)}$ and ``Coupling Coefficient'' $c\ij$~as follows
\begin{eqnarray} \label{eq1}
\hat{\u}_{j|i} = \W_{ij}\u_i \quad \text{ and } \quad
\s_j = \sum_{i=1}^{\NPC} c_{ij}\hat{\u}_{j|i},
\end{eqnarray}
where $\hat{\u}_{j|i}$ is the prediction vector indicating how much the PC $i$~contributes to the CC $j$, and scalar $c_{ij}$ is a coupling coefficient which links predictions of the PC $i$ to the CC $j$. Note that, vector $\s_j$ is a weighted sum of all the PC predictions for the CC $j$. Also we note that, $\NIPC$ denotes the dimension of each capsule in the PC layer while $\NICC$ denotes the dimension of each capsule in the CC layer. We also note that, one can introduce $L \geq 1$ number of intermediate capsule layers located between the PC and CC layers each with $\NIC$ number of localized capsules. If needed, we use index $l$, for ($2 \leq l \leq L+1$), to refer to the  intermediate capsule layers.

Coupling coefficients are trained during the routing process such that for each capsule $j$ in the CC layer, we have $\sum_{i=1}^{\NPC} c_{ij}=1$.  Finally, the vector output for the CC $j$ is obtained through non-linear squashing operator given~by
\begin{equation}
\label{eq2}
\v_j= \frac{\|\s_j\|^2}{1+\|\s_j\|^2}\frac{\s_j}{\|\s_j\|}.
\end{equation}
In each iteration, coupling coefficient $c_{ij}$ is updated by agreement (using dot product) between vector $\v_j$,  which is the output of the CC $j$, and vector $\hat{\u}_{j|i}$, which is the prediction vector of the $i^{\text{th}}$ PC. Length of vector $\v\j$ associated with the CC $j$, for ($1 \leq j \leq \NCC$), indicates the presence of an object represented by  class $j$ in the~input.

Let us further elaborate on the CapsNet structure introduced above, in terms of the example shown in Fig.~\ref{fig:arch}. Here, we have $\NIPC=8$ dimensional PCs, i.e., each capsule entity in the PC layer encodes eight different types of information. Besides, the PC layer consists of $32$ component capsules (cubes) each of which consisting of $36$ capsules, shown by ``arrows'' in the last component capsule. Therefore, the PC layer in this illustrative example has $\NPC = 36\times 32 = 1152$ individual capsules. Dimension of capsules in the CC layer is $\NICC = 2$ and there are total of $\NCC = 10$ individual capsules in the CC layer. Finally, in terms of our running example, $\v_6$ has the largest magnitude among the CCs considered here, therefore, the CapsNet's output decision will be digit 6. This was a brief overview of the CapsNet architecture, next we will investigate its explanability properties.

\section{CapsNets' Explainability}\label{sec:WTE}
In this section, we look at CapsNet architecture described in the previous section from different angles to see whether or not this revolutionized deep learning model has improved explanability characteristics. As we mentioned previously, CNNs can not preserve spatial relationship between components of an object because, some features will be discarded during the pooling process. CNNs compensate this deficiency by increasing the number of training data. In CapsNets, we have capsules (a group of neurons) instead of neurons (single units). Each capsule performs some internal complicated computations and its output is a vector instead of a scalar value. In particular, we investigate the potential properties of using such vectorized outputs, which could lead to more  explainability.

\subsection{Relevance Path by Agreement}
%
\begin{figure}[t!]
\centering
\includegraphics[width=0.45\textwidth]{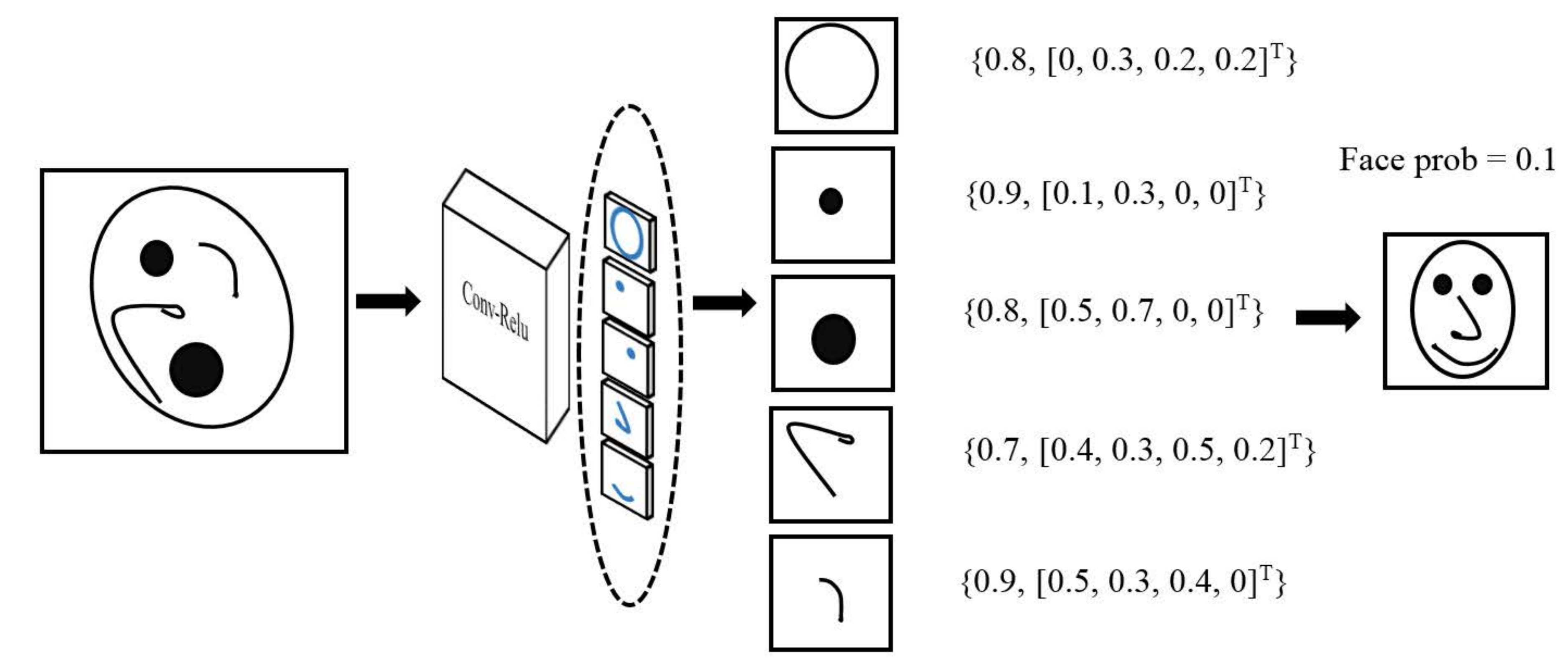}
\caption{\footnotesize Mismatch among instantiation parameter vectors for the face capsule. The first element of each set is the likelihood (probability) of that PC and the second element is the capsule's prediction vector. Although, all the facial components exist with high probability, they disagreement among them explains the resulting low probability for the face capsule. }
\label{fig:face}
\vspace{-.2in}
\end{figure}
%
The vector representation provided by CapsNets is highly informative and can model possible instantiation parameters for components or fragments of an object~\cite{hin:2011}. We argue that this vector output of the capsules can essentially lead to improved explanability of the overall network. In other words, while the length (magnitude) of the output vector $\v\j$ corresponding to capsule $j$ in the CC layer is used to make decisions regarding the input image, the length $\|\u_i\|$ of the output vector $\u\i$ from the of $i^{\text{th}}$ capsule in the PC layer or an intermediate capsule layer can be interpreted as probability of existence of the feature that this capsule has been trained to detect. More specifically, we can assign to each capsule a set consisting of two segments for explanation purposes:
\begin{enumerate}
\item[(i)] Likelihood values which can be used to explain existence probability of the feature that a capsule detects, and;
\item[(ii)] Instantiation parameter vector values which can be used to explain consistency among the layers. In other words, when all capsules of an object are in an appropriate relationship with consistence parameters, the higher level capsule of that object will have a higher likelihood. Therefore, explanations can be provided to describe why the network did detect an object.
\end{enumerate}
For example, Fig.~\ref{fig:face} shows the sets computed based on $5$  intermediate component capsules referring to class face ($j$) in the CC layer. Regarding Item (i), the likelihood part of each of these capsules is relatively high explaining that the input contains all the facial components represented by these 5 component capsules with high probability. However, the network decision is that there is not a face in the input as the likelihood of the face capsule in the CC layer is relatively low. This can be explained based on the non-consistency among the instantiation parameters (Item (ii)).
The output vector of the $i^{\text{th}}$ capsule in the PC layer is multiplied by its weight matrix $\W_{ij}$ corresponding to the $j^{\text{th}}$ capsule in the CC layer, which has been learned through backpropagation algorithm to account for all possible transformations of the instantiation parameters. This multiplication is considered as the vote of capsule $i$ for class $j$ (which is represented by vector $\hat{\u}_{j|i}$). When all related capsules have similar votes for the $j^{\text{th}}$ CC, it means that they agree to each other on presence of object $j$.

CapsNet applies non-linear squashing function on output vectors ($\v_j$) in each iteration. It actually bounds likelihood of these vectors between 0 and 1, which means that it suppresses small vectors and preserves long vectors in the unit length
\begin{equation}
\label{eq3}
\begin{cases}
v_j\approx\|s_j\|s_j\approx 0,& \mbox{if }s_j \mbox{is  small} \\
v_j\approx\frac{s_j}{\|s_j\|}\approx 1,&\mbox{if } s_j \mbox{is  large}
\end{cases}
\end{equation}
Therefore, during agreement iterations, unrelated capsules will become smaller and smaller and the related ones will be remained unchanged. Consequently, introduction of the squashing function results in the coupling coefficients $c_{*j}$ associated with irrelevant capsules to approach zero while coupling coefficient corresponding to the ones responsible for the $j^{\text{th}}$ CC to increase.
Hence, CapsNets intrinsically construct a relevance path (we refers to it as the relevance path by agreement concept) which eliminates the need for a backward process to construct the relevance path. The reason behind the exitance of the relevance path by agreement  is that CapsNet uses dynamic routing instead of common pooling methods.
In the other words, when a group of capsules agree for a parent (higher level component), they construct a part whole relationship which can be considered as a relevance path. For example in face prediction case, facial components (eyes, nose, mouth) in a particular relationship will detect a face as a higher level component (Fig.~\ref{fig:trans}).

\vspace{-.1in}
\section{Experimental Setup} \label{sec:EXP}
\vspace{-.1in}
\begin{table}[t!]	
\caption{\footnotesize CapsNet architecture.}\label{tab3}
\begin{center}
{\tt
\begin{tabular}{|c|}\hline
$28 \times 28 \times 1$\\\hline
Conv with $9 \times 9$ kernels, stride=1, Relu \\\hline
$20 \times 20 \times 256$\\\hline
Conv with $9 \times 9$ kernels, stride=2, 8D Caps \\\hline
$6 \times 6 \times 32 \times 8$\\\hline
Multiply by $W$, Weighted sum, Squashing \\\hline
$10 \times 2 $\\\hline
Decoder part \\\hline
\end{tabular}
}
\end{center}
\vspace{-.25in}
\end{table}
\begin{figure}[t!]
\centering
\includegraphics[width=0.49\textwidth]{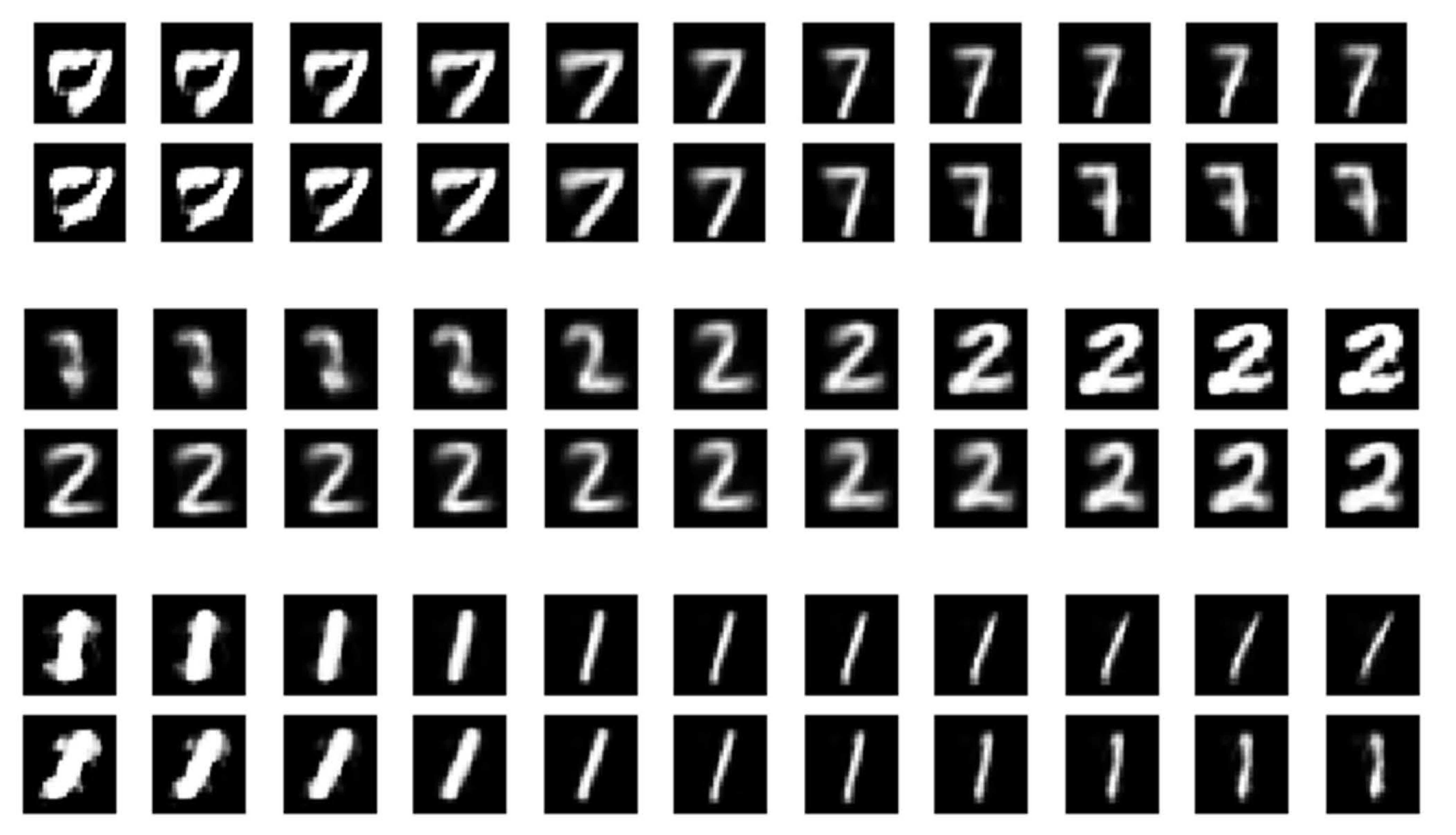}
\caption{\footnotesize Variation of two parameters output of the detected digit capsule within $[-0.25,0.25]$ with step size of $0.1$ for three sample digits. One can then explain the learned features of the CCs as thickness and deformation.}
\label{fig:ex1}
\vspace{-.2in}
\end{figure}
\begin{figure}[t!]
\centering
\includegraphics[width=0.29\textwidth]{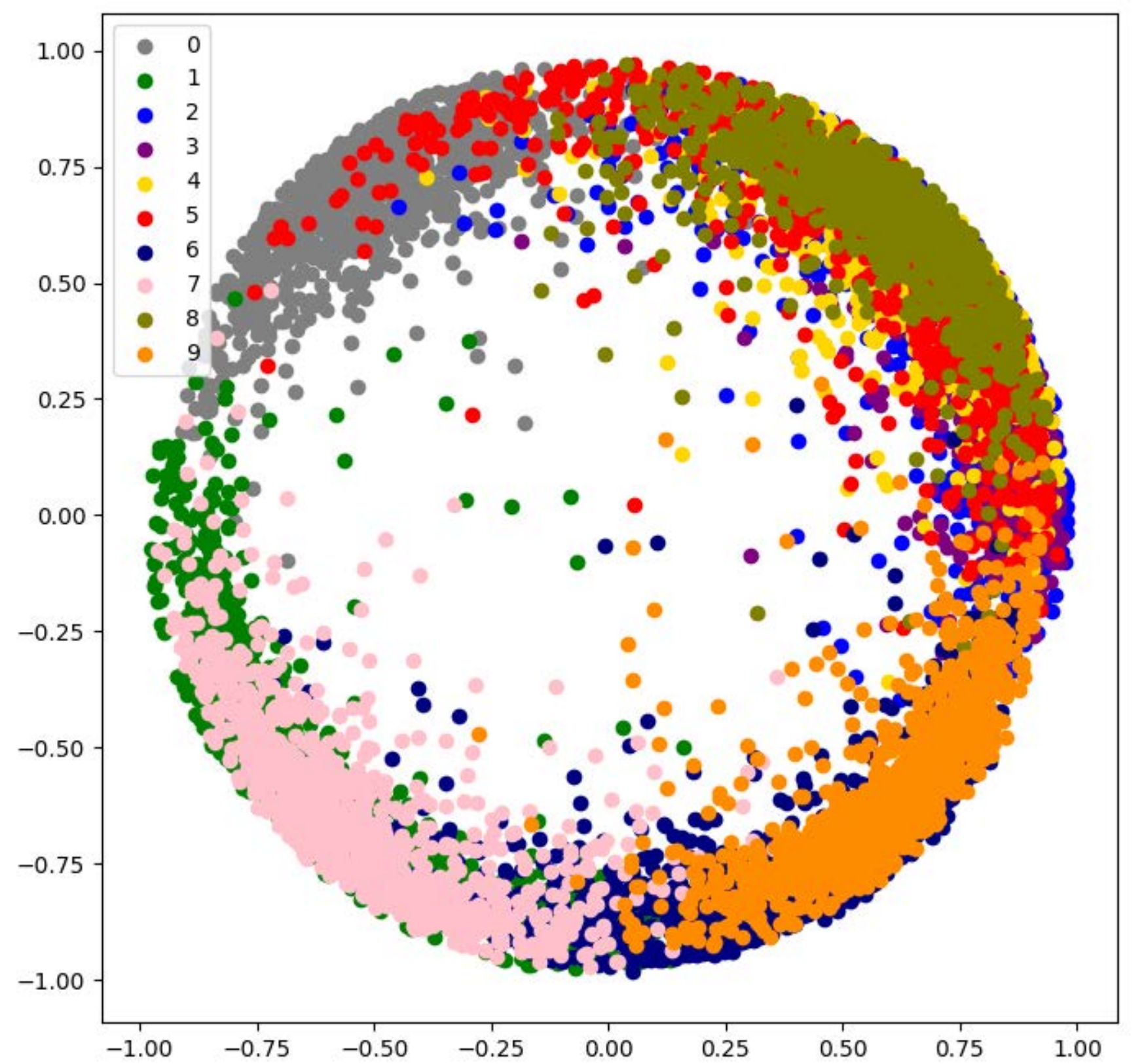}
\caption{\footnotesize Two parameters output vector of detected digit capsules.}
\label{fig:ex3}
\vspace{-.1in}
\end{figure}
\begin{figure}
\centering
\includegraphics[width=0.49\textwidth]{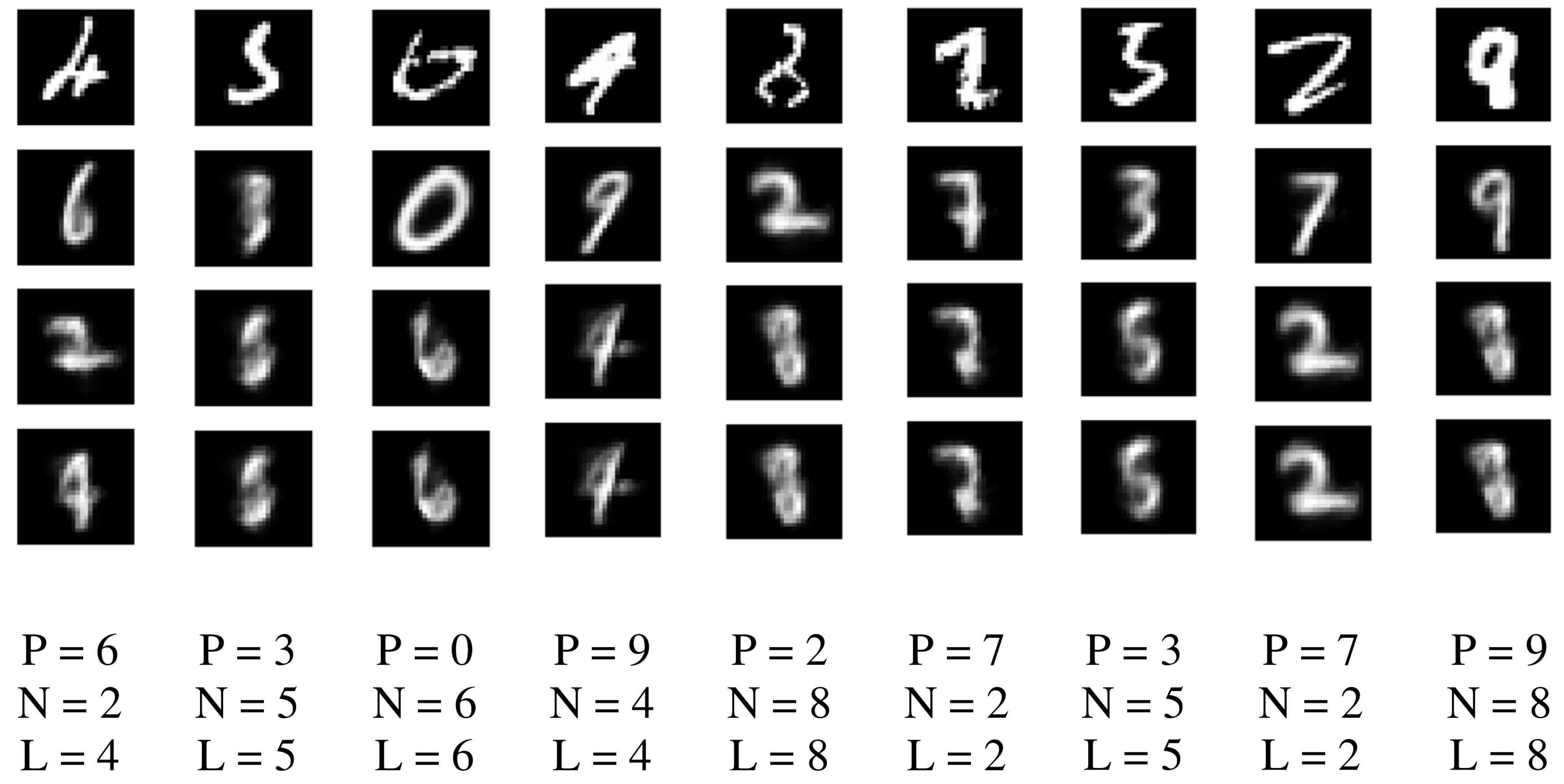}
\vspace{-.2in}
\caption{\footnotesize Misclassified samples: The 1st row represents the input digit; The 2nd row shows the reconstructed digit by the CapsNet's prediction; the 3rd row is the reconstruction based on the capsule with second highest likelihood, and; finally the last row is the reconstructed digit by true label.}
\label{fig:ex2}
\vspace{-.25in}
\end{figure}
In this section, we investigate explanation capabilities of the CapsNets on $28 \times 28$ MNIST dataset~\cite{mnist:1998}. The used architecture is similar to the one that has been presented in~\cite{Caps:2017} as outlined in Table~\ref{tab3}.
After training CapsNet with MNIST training data, in a first experiment we varied each of the two parameters of the CC output vector corresponding to the detected digit and reconstructed it again to find out what is nature of the two plausible features that have been learned (i.e., to explain the learned features at the CC layer). Fig.~\ref{fig:ex1} illustrates the results for three digits. It is observed that by changing the parameters, thickness and shape of digits have been changed simultaneously. Therefore, we can consider them as explanation of the learned features.
Fig.~\ref{fig:ex3} displays the two parameters  of the output vector associated with the detected digit capsule for all the testing dataset. As can be seen, the vector output of different digit capsules significantly overlap with each other, e.g., capsules 1/7; 6/9, and; 3/5 generate close values. Therefore, in misclassified samples, there is a high probability for detecting overlapped digits. Recognizing these failure points is another level of explanation that CapsNet can provide on MNIST dataset.
Now, we found misclassified samples in 1000 testing data and display their output prediction vectors. In all cases, the true class had the second highest likelihood (magnitude) except one instance in which the third highest likelihood was corresponding to the true label. Moreover, we can usually see a high decreasing magnitude between third and forth positions. Therefore, one can use the likelihood values and present the second and third highest capsules as alternative solutions and explanations. Fig.~\ref{fig:ex2} shows all misclassified samples in 1000 testing data. We reconstructed input digits by prediction and true label outputs. As we see, each input digit in the first row is similar to both reconstructed digits bellow it, which explains why the model has been failed in these cases.

\vspace{-.2in}
\section{Conclusion}  \label{sec:con}
\vspace{-.1in}
In this paper, we represented the necessity of explainability in deep neural networks especially in critical decisions where a single incorrect decision is even unacceptable. Previous explainability methods try to find and visualize the most relevant pixels or neurons by adding an extra explanation phase. In this work, we illustrated potential intrinsic explainability properties of Capsule network,  by analyzing its behavior and structure.


\end{document}